%% file: AAAI.tex
\documentclass[letterpaper]{article} 
\usepackage{aaai24}  
\usepackage{times}  
\usepackage{helvet}  
\usepackage{courier}  
\usepackage[hyphens]{url}  
\usepackage{graphicx} 
\usepackage{natbib}  
\usepackage{caption} 
\usepackage{algorithm}
\usepackage{algorithmic}

\usepackage{amsfonts}
\usepackage{amsmath}
\usepackage{booktabs}
\usepackage{multirow}
\usepackage{subfig}
\usepackage{pifont}
\usepackage{array}
\newcolumntype{L}[1]{>{\raggedright\let\newline\\\arraybackslash\hspace{0pt}}m{#1}}
\newcolumntype{C}[1]{>{\centering\let\newline\\\arraybackslash\hspace{0pt}}m{#1}}
\newcolumntype{R}[1]{>{\raggedleft\let\newline\\\arraybackslash\hspace{0pt}}m{#1}}
\newcommand{\tabincell}[2]{\begin{tabular}{@{}#1@{}}#2\end{tabular}}

\usepackage{newfloat}
\usepackage{listings}
\DeclareCaptionStyle{ruled}{labelfont=normalfont,labelsep=colon,strut=off} 
\floatstyle{ruled}
\newfloat{listing}{tb}{lst}{}
\floatname{listing}{Listing}
\title{A Dual-Way Enhanced Framework from Text Matching Point of View for Multimodal Entity Linking}
\author{
    Shezheng Song\textsuperscript{\rm 1, 2},
    Shan Zhao\thanks{Corresponding Author}\textsuperscript{\rm 2},
    Chengyu Wang\textsuperscript{\rm 1},
    Tianwei Yan\textsuperscript{\rm 1}\\
    Shasha Li$^*$\textsuperscript{\rm 1},
    Xiaoguang Mao\textsuperscript{\rm 1},
    Meng Wang\textsuperscript{\rm 2}
}
\affiliations{
    \textsuperscript{\rm 1}National University of Defense Technology\\
    \textsuperscript{\rm 2}Hefei University of Technology\\

    betterszsong@gmail.com, zhaoshan@hfut.edu.cn, shashali@nudt.edu.cn
}

\usepackage{bibentry}

\begin{document}

\maketitle

\begin{abstract}
    Multimodal Entity Linking (MEL) aims at linking ambiguous mentions with multimodal information to entity in Knowledge Graph (KG) such as Wikipedia, which plays a key role in many applications. However, existing methods suffer from shortcomings, including modality impurity such as noise in raw image and ambiguous textual entity representation, which puts obstacles to MEL. 
    We formulate multimodal entity linking as a neural text matching problem where each multimodal information (text and image) is treated as a query, and the model learns the mapping from each query to the relevant entity from candidate entities. This paper introduces a dual-way enhanced (DWE) framework for MEL: (1) our model refines queries with multimodal data and addresses semantic gaps using cross-modal enhancers between text and image information. Besides, DWE innovatively leverages fine-grained image attributes, including facial characteristic and scene feature, to enhance and refine visual features. (2)By using Wikipedia descriptions, DWE enriches entity semantics and obtains more comprehensive textual representation, which reduces between textual representation and the entities in KG.
    Extensive experiments on three public benchmarks demonstrate that our method achieves state-of-the-art (SOTA) performance, indicating the superiority of our model. The code is released on \url{https://github.com/season1blue/DWE}.
\end{abstract}

\section{Introduction}
\begin{figure}[pbt]
        \centering
        \includegraphics[width=\linewidth]{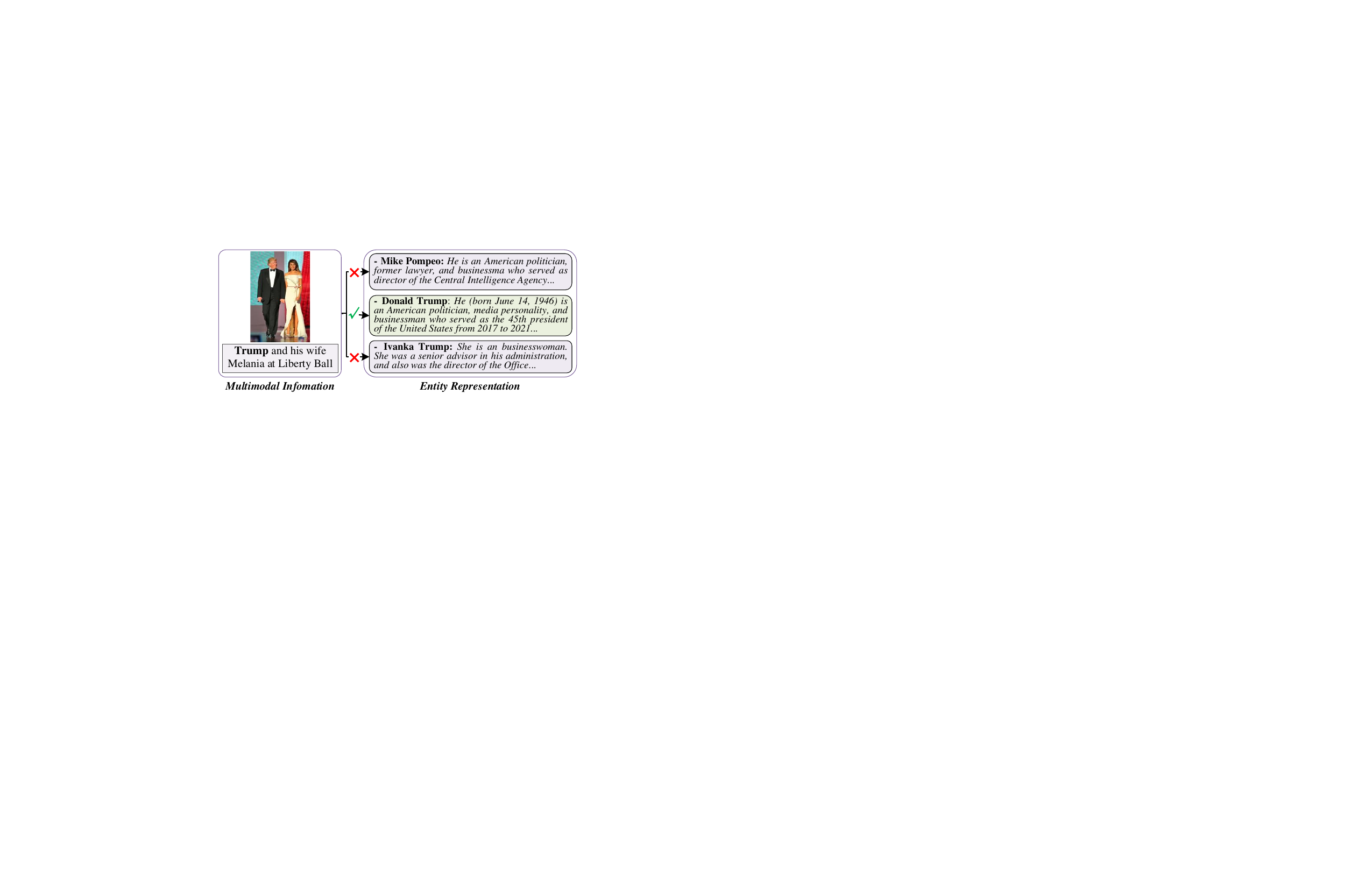}
        \caption{Example of entity linking for the mention \textit{Trump}.}
        \label{fig:intro1}
    \end{figure}

    Entity linking (EL) \cite{el1, el2} has attracted increasing attention in the natural language processing (NLP) domain, which aims at linking ambiguous mentions to the referent unambiguous entities in a given knowledge base (KB) \cite{ kb1}. It is crucial in many information retrieval applications, such as information extraction \cite{ ie3}, question answering \cite{intro_qa1}, and Web query \cite{intro_sq3}. However, the ambiguity in mentions poses great challenges to this task. Hence,  Multimodal Entity Linking (MEL) has been introduced recently
    \cite{mel_method1, mel_method2}, where multimodal information (e.g., visual information) is used to disambiguate the mention.
    Figure \ref{fig:intro1} shows an example of MEL, wherein the mention \textit{Trump} is associated with the text \textit{``Trump and his wife Melania at the Liberty Ball"}. It is challenging to distinguish the entities because the short text is ambiguous, and several entities are related to \textit{Trump}. Thus, image is required for further disambiguation. 
    Dozens of MEL studies have been conducted in the past few years and achieved promising performances, where delicate interaction and fusion mechanisms are designed for encoding the image and text features \cite{baseline_mel,baseline_dataset,hybrid}. Nonetheless, existing approaches still struggle to fully exploit the potential of the feature sources from two distinct information perspectives, which may hinder further task development.
    
    \textit{ \textbf{Modality impurity:}}
In MEL task, one of the challenges lies in the fact that the image modality often contains less information and more redundant features when compared to the text modality.  Indeed, not all the visual sources play positive roles. As revealed by \citeauthor{vempala2019categorizing}, as high as 33.8\% of visual information serves no context or even noise in task.
As shown in Figure \ref{fig:intro1},  only the left person is relevant while the right person is noise, causing redundant visual elements and low information density for the entity. 
However, for the existing MEL works, they may
suffer from ineffective and insufficient modeling of visual pivot features. The majority of MEL models perform vision-language grounding by simultaneously considering the entire image and text \cite{zhang2019neural,fei2023scene}.  Unfortunately, such coarse-grained representation learning often leads to information mismatches and compromises the subtle semantic correlations between vision and language. 
We argue that a fine-grained feature complementing between text features and visual information is needed.

\textbf{\textit{Ambiguous Entity Representation:}}
 Wang et al. \shortcite{baseline_mel} and Zhou et al. \shortcite{baseline_dataset}  concatenate the main properties of an entity into a piece of text, which is regarded as a representation of the entity in KG. \citeauthor{wikiperson}\shortcite{wikiperson} manually collect short text to represent entity, e.g., ``Bahador Abb: Iranian footballer". However, the set of properties or short text are insufficient to represent entities. Different people may have the same properties. For example, the entity \textit{Donald Trump} and \textit{Mike Pompeo} both have the properties of \textit{``Sex: male. Religion: Presbyterianism. Occupation: businessperson, politician. Languages: English. work location: Washington, D.C."} With so similar entity representation, it is a great challenge to link entity even with an ideal multimodal method. Thus, more distinctive and representative entity representation is required for further disambiguation. Fortunately,  Wikipedia description offers a promising solution, which has been shown to enrich the semantics of raw data, especially in low-resources NLP tasks \cite{chen2022few}.
 
\begin{figure*}[h]
        \centering
        \includegraphics[width=\textwidth]{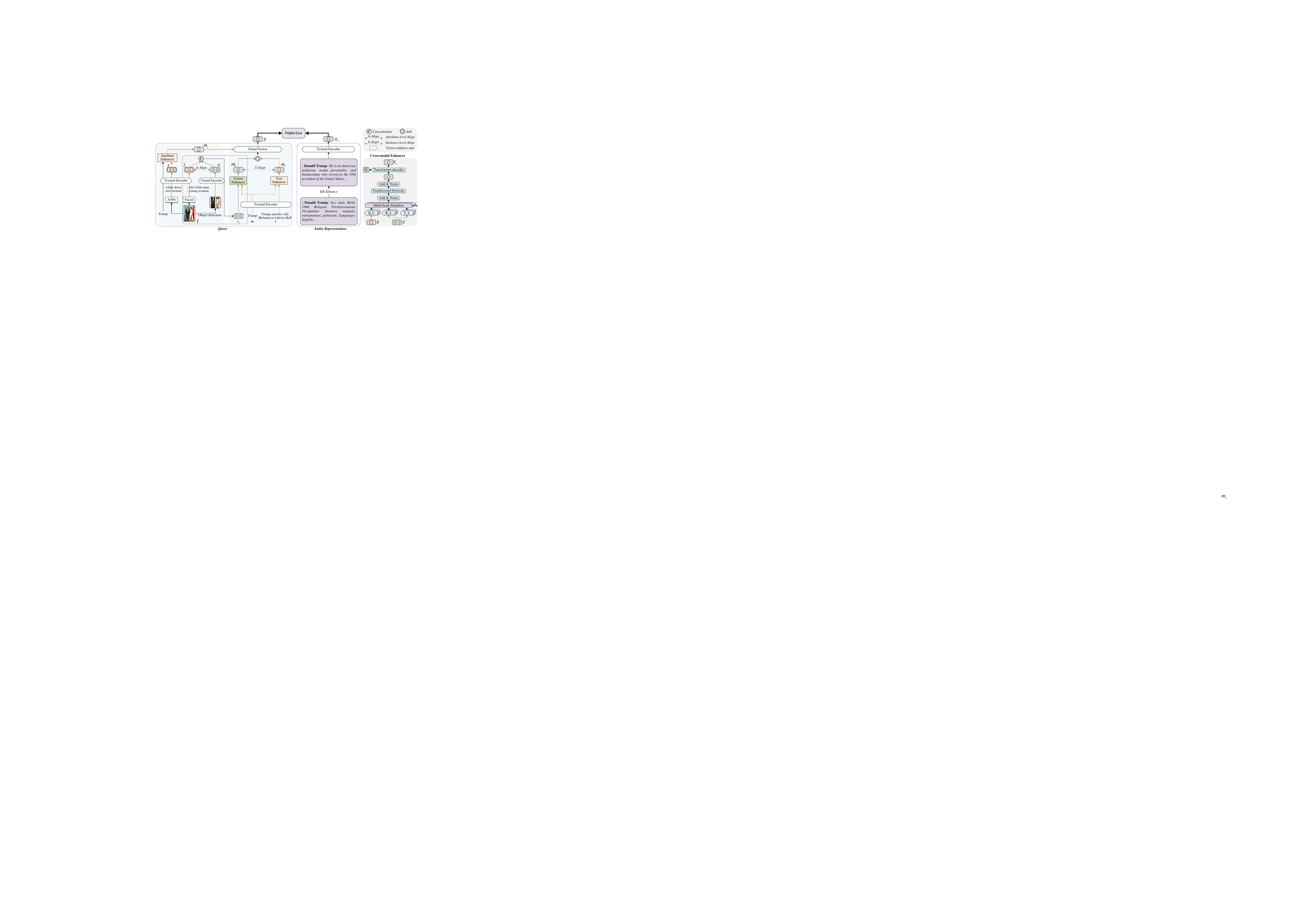}
        \caption{
            Overview of DWE.
            Input consists of image $I$, text $t$, and mention $m$. Object detection is applied to extract object feature $d_i$ from image. Facial feature $f$ and scene feature $s$ (i.e. Adjective-Noun Pairs(ANPs))  is retrieved from image. The fusion feature $g$ is combination of image-enhanced feature $m_v$, text-enhanced feature $m_t$ and attribute-enhanced feature $m_s$.
        }
        \label{fig:model}
    \end{figure*}

To tackle the above issues, in this work, we propose a novel framework  to improve  MEL.
Inspired by neural text matching task, we view each mention as a query and attempt to learn the mapping from each mention to the relevant entity. To achieve this, we design a dual-way enhanced \textbf{(DWE)} framework:  enhancing  query with refined multimodal information;  enriching the semantics of  entity by  Wikipedia description as shown in  Figure  \ref{fig:model}. 
Specifically, \textbf{(1)} for  the enhancing  query (e.g., mention), DWE employs a pretrained visual encoder and visual tools to obtain the image  representation  and  fine-grained visual attributes, then purify it
into the learnable visual characteristics. The visual characteristics and text are regarded as two types of
refined information, which are separately employed to enhance the mention.  
   Considering the semantic gap between vision and text in the language model, we design
   three enhanced units based on the cross-modal enhancer
to bridge the semantic gap between (i) facial feature and object visual feature (ii) mention and vision-enhanced feature and (iii)mention and text-enhanced feature. By introducing the cross-modal alignment, our model makes use of the semantics of text and objective image information.
   \textbf{ (2)}
    On the other hand, we utilize a knowledge base (i.e. Wikipedia) to enrich the semantics of entity representation. This process reduces the disparity between textual representation and the entities in the knowledge graph, resulting in a more comprehensive textual representation. To sum up, we enhance entity representations for 17805, 17291, and 57007 entities in the Richpedia, WikiMEL, and Wikidiverse datasets, respectively.

    Extensive experiments show that DWE pushes the state-of-the-art \textbf{(SOTA)} by a large margin, indicating the efficiency of DWE.
    Our contributions are as follows:
    

    \begin{itemize}
        \item We explore Multimodal Entity Linking in a neural matching formulation. We  design a dual-way enhanced framework: query is enhanced by refined multimodal information and we utilize the Wikipedia to enrich the semantics of entity representation.

        \item We innovatively introduce fine-grained image attributes (such as facial characteristics and scene feature) which are leveraged to enhance and refine the visual features and take cross-modal alignment to bridge the semantic gap between textual and visual features. 
        
        \item Experimental results on three benchmark datasets Richpedia, WikiMEL and Wikidiverse demonstrate the superiority of our proposed model over the SOTA method.
    \end{itemize}


\section{Related Work}
    MEL has been widely studied in recent years, which leverages the associated multimodal information to better link mention to entity in KG. Previous studies mainly focus on the following two aspects:

    \textbf{Multimodal Information Refinement}
    Recently, dozens of works have focused on utilizing multimodal information for disambiguation. Wang et al. \shortcite{baseline_mel} propose to use gated multimodal feature fusion and contrastive learning for MEL. \citeauthor{wikidiverse} \shortcite{wikidiverse} use ResNet\cite{resnet} as a visual encoder and BERT\cite{BERT} as a textual encoder to extract the joint feature for MEL. However, these studies~\cite{zhao2023mcl, zhao2021enhancing, zhao2021dynamic} ignore the noise of the image, i.e., some objects in the image are irrelevant to the mention. \citeauthor{hybrid} \shortcite{hybrid} notice the visual and textual noise and take image segmentation and attention mechanism as refinement method, but ignore the global relationship between text and image. \citeauthor{ita} \shortcite{ita} also realize the noise, but they pay too much attention to extracting text such as OCR text or caption from raw image. To conclude, these work~\cite{ma2023using, ma2022joint} can not fully use multimodal information and filter the irrelevant part (i.e., noise) of raw image effectively. 
    
    \textbf{Ambiguous Entity Representation}
    It is still controversy over how to represent the entity in KG, so there are different methods of entity representation. \citeauthor{tweet} \shortcite{tweet} propose a framework for automatically building the MEL dataset from Twitter\cite{twitter}, taking profile on Twitter as the user entity representation. However, the profile is edited by user and can not fully represent the user entity. \citeauthor{baseline_dataset} \shortcite{baseline_dataset} propose three MEL datasets, built from Weibo, Wikipedia, and Richpedia information, collecting the property of entity as representation, such as ``Donald Trump: male, 1946, politician" and using Wikidata \cite{Wikidata2} and Richpedia \cite{Richpedia} as the corresponding KGs. Similarly, Wikiperson\cite{wikiperson} dataset takes a manually organized short text as entity representation such as ``Bahador Abdi: Iranian footballer".
    However, the entity representation of the above datasets is ambiguous, making it hard to distinguish the MEL target (entity) and affecting the plausibility of performance evaluation.
    Our work differs from the above papers in that we use the textual Wikipedia description as entity representation to guide the learning of multimodal feature, which are proven to be more efficient.

\section{Methodology}
    
    \subsection{Notation and Task Definition}
        Let $X = \{x_i\}_{i=1}^N $ be a set of $N$ input multimodal samples, with corresponding entities set $Y = \{e_i\}_{i=1}^M$ in a KG. Each input is composed of three parts: $x = \{x_m; x_t; x_v\}$, where $x_t$ denotes a text consisting of a sequence of tokens with length $l_t$, $x_v$ is an image associated with the text, and $x_m$ is a set of mentions in text $x_t$.
        
        A KG is composed of a set of entities $Y = \{e_i\}_{i=1}^M$. Each entity $e_i$ is described by a sequence of entity representations $e_i= u_1, u_2... u_{w}$ (i.e., textual Wikipedia description about the entity). $w$ is the token length of entity representation.
        The MEL task aims to link the ambiguous mention (i.e. query) to the entity in KG by calculating the similarity between the joint mention feature and textual entity representation. The entity with the highest similarity is regarded as the corresponding linking entity. In this process, textual and visual information is used to disambiguate and further clarify semantics. Finally, the KG entity $y$ linked to mention is selected from $\lambda$ candidate entities $C_e= \{e_i\}_{i=1}^\lambda$. 
        It can be formulated as follows:
    	\begin{equation}
            \footnotesize
    		y = \mathop{argmax} \limits_{\forall e \in C_e} \Gamma(\Phi(x), \Psi(e))
    	\end{equation}
        where $\Phi$ represents the multimodal feature function, $\Psi$ represents the KG entity representation feature function, and $\Gamma()$ produces a similarity score between the multimodal representation and entity representation in KG. $y$ is the predicted linking entity.

        In evaluation, we take cosine similarity to measure the correlation between the joint multimodal feature $g$ and the candidate entities $C_e$. The entity with the highest similarity to $g$ is regarded as the corresponding linking entity. For the example shown in Figure \ref{fig:intro1}, The similarities between the mention and entities are 0.47, 0.81 and 0.13. Therefore, candidate entity \textit{Donald Trump} is chosen as the linked entity.

    \subsection{Preliminary}
        \textbf{Knowledge base Enhanced Entity Representation.}
        Previous works \cite{baseline_mel, baseline_dataset} tend to concatenate the main properties of an entity into a piece of text and regard the properties as textual entity representation. However, a set of properties is not clear enough to distinguish the linking entity from candidate entities. Different entity may be the same in some fields. As shown in the entity representation in Figure \ref{fig:model}, the properties of \textit{Mike Pompeo} is similar with \textit{Donald Trump} in \textbf{previous} representation. Thus it is challenging to distinguish the two people with nearly the same properties.

        Therefore, we leverage Wikipedia to get \textbf{enhanced} representation, with the textual Wikipedia description from the text on Wikipedia pages. In summary, 17805, 17391, and 57007 entity representation are enhanced on Richpedia, WikiMEL, and Wikidiverse datasets respectively. 

        \textbf{Textual feature extraction.}
        Given a sentence $x_t$, mention $x_m$ and Wikipedia description of 
        entity  $x_e$, we follow CLIP \cite{clip_model} to tokenize it into a sequence of word embeddings. Then the special tokens \textit{startoftext} and \textit{endoftext} are added at the beginning and end positions of word embeddings. As a result, with $N$ sentences and $N_e$ candidate entities, we feed  sentence representation  $t \in \mathbb{R}^{N \times d}$,  mention representation $m \in \mathbb{R}^{N \times d}$ and  enhanced entity representation $g_e \in \mathbb{R}^{N_e \times d}$ into model.

    \subsection{Multimodal Information Enhanced Query }

We first present a  general architecture: cross-modal enhancer, as shown on the right of Figure \ref{fig:model}. Then we build three enhanced units based on this
architecture to capture dense interaction over multimodal information,  namely a text-enhanced unit, an attribute-enhanced
and a vision-enhanced unit.

\paragraph{Cross-modal enhancer.} 
Given two types of  features, $X$ and $Y$, the  enhanced features  is calculated as:
 \begin{equation}
        \footnotesize
            E_n = E_n(X, Y)
        \end{equation}
Specifically,    the calculation of multi-head attention for hidden state $h_t$ is  first performed on the hybrid key and value:
        \begin{equation}
        \footnotesize
            h_t = Softmax(\frac{(W^QX)(W^KY)^T}{\sqrt{d}})(W^VY) 
        \end{equation}
    Then we utilize the transformer decoder for decode  hidden state $h_t$ with the learnable textual query $Q_v \in \mathbb{R}^{N_v\times d}$:
        \begin{equation}
        \footnotesize
            E_n = Decoder(h_t, Q_v)
        \end{equation}
    In detail, we formulate the attention of transformer decoders in Decoder as:
        \begin{equation}
        \footnotesize
            E_n = Softmax(\frac{(W^QQ_v)(W^Kh_t)^T}{\sqrt{d}})(W^Vh_t) 
        \end{equation}
     where $W^Q \in \mathbb{R}^{d\times d_q}, W^K \in \mathbb{R}^{d\times d_k}, W^V \in \mathbb{R}^{d\times d_v}$ are randomly initialized projection matrices. We set $d_q=d_k=d_v=d/h$. $h$ is the number of heads of attention layer.

\paragraph{Text-enhanced unit.} 
The textual-enhanced  is designed to model sentence-to-mention correlations.
It takes two groups of feature 
sentence representation  $t \in \mathbb{R}^{N \times d}$,  mention representation $m \in \mathbb{R}^{N \times d}$ as inputs.
Next, the text-enhanced unit models pairwise relationship between each paired sample ($m,t$) and finally outputs enhanced mention features:
  \begin{equation}
        \footnotesize
            m_t = E_n(m, t)
        \end{equation}

\paragraph{Attribute-enhanced unit.} 
    Sentibank\cite{anpextractor} is applied to extract adjective-noun pairs (ANPs) $x_s$ such as ``nice clouds, white dress, happy man.." from image, which is relevant to the attributes of scene feature. Similarly, we integrate and encode $x_s$ to get attribute feature $s$.
    We apply cross-modal enhancer to get the attribute-enhanced mention feature $m_s$.
        \begin{equation}
        \footnotesize
            m_s = E_n(m, s)
        \end{equation}
    where $s \in \mathbb{R}^{n_s\times d}$ and $m_s \in \mathbb{R}^{1 \times d}$, $d$ indicates the hidden dimension and $n_s$ is the number of adjective-noun pairs.

\paragraph{Vision-enhanced unit.} 

    Multimodal information in the MEL task generally consists of textual and visual information. Text is artificial, and thus its semantics is intensive. In contrast, visual information tends to be a photo of objects or scenes. There is inevitably some redundant information in the raw image. For an image with multiple mentions, only some interested objects in the image are related to one mention, while the rest part is related to the other mention. Some objects or scenes are still irrelevant and disturbing, even for an image with a mention. Therefore, refining the raw image by detecting and extracting the related visual object is necessary.
    As shown in Figure \ref{fig:model}, the text is \textit{Trump and his wife Melania at the Liberty Ball}, and the mention is \textit{Trump}. Only the local region of the left person contributes to linking mention \textit{Trump} and entity \textit{Donald Trump} in Wikidata, while the rest of the image is of less significance.
    
    An image $I$ can be decomposed into $l$ objects in local regions. Each mention only describes a part of the image. Therefore, we take an object detector \textbf{\textit{D}}\cite{pixellib} to identify all possible objects in the image. The object visual feature $d$ is extracted through the image encoder of CLIP.
        \begin{align}
        \footnotesize
            x_d & = \{x_d^i \}_{i=1}^{l} = D(I) \\
            d & = clip(x_d)
        \end{align}
    where $d = \{d_{i}\}_{i=1}^{l}$ and $l$ is the number of the detected objects in the image. 

    Deepface\footnote{https://github.com/serengil/deepface} is applied to extract facial appearance $x_f$ such as gender, race, age and etc,  which are relevant to the attributes of facial feature in the image. $x_f$ is integrated into prompt sentence such as ``Trump, gender: male, race: white, age: 50..." and encoded through image encoder of CLIP to get facial feature $f= \{f_{i}\}_{i=1}^{l}$.
    Due to $f_i$ and $d_i$ is facial feature and local visual feature of i-th object of image, the visual feature of i-th object $v_i$ is enhanced by $v_i=[d_i, f_i]$, where refined visual feature $v = \{v_i\}_{i=1}^{l}\in \mathbb{R}^{l \times d}$ and [,] denotes concatenation. 
    
    Only some objects in the image are related to the mention. Thus, we utilize the visual enhancer to determine related visual feature and obtain vision-enhanced mention feature $m_v$.
        \begin{equation}
        \footnotesize
            m_v = E_n(m, v)
        \end{equation}


\input{Tables/main}

        \subsection{Gated Fusion and Training Loss}
        we get the feature of mention $m$, text-enhanced mention $m_t$, vision-enhanced mention $m_v$ and attribute-enhanced mention $m_s$ and apply gated fusion to restrain noise in $m_s$. 
    	\begin{align}
            \footnotesize
    		g_m &= m + m_t + m_v \\
                g &= (g_m + \alpha (m_s - g_m)) W_g  + b_g
    	\end{align}
        where $W_g \in \mathbb{R}^{2d \times d_{m}}$, $d$ is hidden dimension, $d_{m}$ is the dimension of joint feature $g$. $b_g$ and $\alpha$ is learnable parameters.

        During training, we attempt to maximize the similarity between the mention and its corresponding entity (positive sample $\theta_+$) in KG and minimize the similarity between the mention and other entities (negative sample $\theta_-$).
        Thus we take triplet loss\cite{triplet} to maximize the distance between multimodal feature $g$ and negative samples while minimizing the distance between $g$ and positive samples. The triplet loss $\mathcal{L}_t$ is defined as:
    	\begin{equation}
            \footnotesize
    		\mathcal{L}_t = max( \Gamma(g, \theta_+) - \Gamma(g, \theta_-) + \varepsilon, 0)
    	\end{equation}
        $\Gamma$ is cosine similarity between features. $\varepsilon$ denotes margin, a hyperparameter for learning.
        
        \paragraph{Cross-modal alignment.}
        To bridge the semantic gap between features, we design (1)attribute-level alignment for facial feature $f$ and object visual feature $d$, and (2)sentence-level alignment for text-enhanced feature $m_t$ and vision-enhanced feature $m_v$
            \begin{equation}
            \footnotesize
    		\mathcal{L}_c = L_{m}(f, d) + L_{m}(m_t, m_v) 
    	\end{equation}
        where $L_{m}$ is Mean-Shifted Contrastive Loss (MSC) \cite{loss}. 
        
        The overall loss is the combination of triplet loss and crossmodal alignment loss. $\beta$ is the coefficient factor for balancing the losses of two tasks: $\mathcal{L} = \mathcal{L}_t + \beta \cdot \mathcal{L}_c $.

\section{Experiments}
    
    \subsection{Datasets and Settings}
        We take three public datasets: Richpedia\cite{Richpedia}, WikiMEL\cite{baseline_dataset}, and Wikidiverse\cite{wikidiverse} as benchmark. Table \ref{tab:statistics} shows the statistics of datasets. Besides, we apply enhanced entity representation on three datasets and reproduce several baselines for comparison. 
            \begin{table}[tb]
              \centering
                \begin{tabular}{lcccc}
                \toprule
                Dataset & Sample & Entity & Mention & Text length \\
                \midrule
                Richpedia & 17805 & 17804 & 18752 & 13.6 \\
                WikiMEL & 18880 & 17391 & 25846 & 8.2 \\
                Wikidiverse & 13765 & 57007 & 16097 & 10.1 \\
                \bottomrule
                \end{tabular}%
              \caption{The statistics of three multimodal datasets}
              \label{tab:statistics}
            \end{table}%

	\subsubsection{Metrics}
    	We use Top-k accuracy as the metric:
    	\begin{equation}
            \footnotesize
                Acc_{top\text{-}k} = \frac{1}{N}\sum_{i=1}^{N}\eta \{ I(cos(g, gt), cos(g, C_e)) \leq k \}
    	\end{equation}
            where $N$ represents the total number of samples, and $\eta$ is the indicator function. When the receiving condition is satisfied, $\eta$ is set to 1, and 0 otherwise. $gt$ is ground truth entity while $C_e$ is a set of candidate entities. $cos$ means cosine similarity function. $I$ is function to calculate the rank of similarity between joint feature $g$ and ground truth $gt$ among a set of candidate entities $C_e$.

	\subsubsection{Hyperparameters and Training Details}
            During the experiments, the dimension of text representation $t$ and visual representation $v$ are set to 512, the dimension of related visual object feature $R$ is set to 768, and the heads of Multi-head attention is set to 8.  The dropout, training epoch, evaluating steps, weight decay, and triplet loss interval are set to 0.4, 300, 2000, 0.001, and 0.5, respectively. We optimize the parameters using AdamW \cite{adam} optimizer with a batch size of 64, the learning rate of $5 \times 10^{-5}$ and coefficient factor $\beta$ of 0.5.
            All the experiments is processed on RTX3090 and Pytorch 2.0.
            
        \subsubsection{Candidate Retrieval}
            Following previous work\cite{baseline_dataset, wikidiverse}, we set the number $\lambda$ of candidate entities to 100. The strategy of candidate entity selection varies on different datasets. In Richpedia and WikiMEL, we apply \textit{fuzz}\footnote{https://github.com/seatgeek/fuzzywuzzy} to search 100 candidate entities whose name is similar with mention. In Wikidiverse, the dataset provides 10 similar entities. We first divide entities into different collections according to entity types, such as person, location, and so on. According to mention type, we use \textit{fuzz} to search 90 entities with names similar to mention.

        \subsubsection{Negative Samples}
            We employ two negative sampling strategies for triplet loss: hard negatives and in-batch negatives. The hard negatives are the candidate entities retrieved in the Candidate Retrieval step except for the gold entity. The in-batch negatives are gold entities of other $B-1$ mentions in the batch, where $B$ is batch size.

    \subsection{Baseline}
        we conduct comparative experiments to evaluate the effectiveness of our model on the previous entity representation (combination of properties) for comparison. Furthermore, previous entity is ambiguous and is not representative for entity. Thus, we propose the enhanced entity representation (textual Wikipedia description) and conduct extensive experiments on it.

        (1) \textit{BLINK} \cite{BLINK} is an BERT-based EL model with a two-stage zero-shot linking.
        (2) \textit{BERT} \cite{BERT} stacks several layers of transformer to encode each token in text.
        (3) \textit{ARNN} \cite{ARNN} utilizes Attention-RNN to predict associations with candidate entity textual features.
        (4) \textit{DZMNED} \cite{DZMNED} takes concatenated multimodal attention mechanism to fuse visual, textual, and character features of mentions.
        (5) \textit{JMEL} \cite{jmel} utilizes fully connected layers to project the visual and textual features into an implicit joint space.
        (6) \textit{MEL-HI} \cite{MELHI} uses multiple attention to get richer information and refine the negative impact of noisy image.
        (7) \textit{HieCoAtt} \cite{HieCoAtt} is a multimodal fusion mechanism, using alternating co-attention and three textual levels (tokens, phrases, and sentences) to calculate co-attention maps.
        (8) \textit{GHMFC} \cite{baseline_mel} is a advanced baseline proposed by Wang et.al \cite{baseline_mel}. It takes the gated multimodal fusion and novel attention mechanism to link multimodal entities.
        (9) \textit{MMEL} \cite{MMEL} is a joint feature extraction module to learn the representations of context and entity candidates, from both the visual and textual perspectives. 
        (10) \textit{CLIP-text} \cite{clip_model} only use textual information such as text and mention in the dataset. It mainly focuses on the ability to understand textual relationship among text, mention, and entity representation.
        (11) \textit{CLIP} \cite{clip_model} take both textual and visual features into consideration. The model concatenates multimodal features and calculates the similarity between features and ground-truth.

\input{Tables/ablation}

    \subsection{Main Results}
        On Richpedia\cite{Richpedia}, WikiMEL\cite{baseline_dataset} and Wikidiverse\cite{wikidiverse} datasets, we compare the proposed model with several competing approaches and show the results in Table \ref{tab:experiment} under different entity representation. 
        Under previous entity representation, our method could reach 67.6\%, 44.7\%, and 47.5\% on three datasets.
        Besides, under enhanced entity representation, our method also achieves state-of-the-art (SOTA) performance by obtaining Top-1 accuracy of 72.5\%, 72.8\%, and 51.2\%, respectively, which demonstrates the superiority over the previous state-of-the-art method.        
        Apparently, with pre-trained on massive multimodal data, the text and image encoder of clip facilitates the understanding of complicated multimodal information. For the gated fusion, it benefits models by restraining noise while using information. Besides, the enhancer for text and vision contributes to fusion between features.

        \textbf{Enhanced Entity Representation}. Previous works \cite{baseline_dataset,baseline_mel} consider the properties as entity representation, which is ambiguous and limited. we take textual Wikipedia description as enhanced representation. Results show that our method is not only useful for previous entity representation but also the enhanced representation. Besides, on CLIP and our method, the improvement over different representations shows that the enhanced representation is helpful for disambiguation. In detail, on Richpedia, by using enhanced representation, CLIP obtain the 3.5\% improvement and our method obtains 4.9\% improvement. The boost proves that the enhanced entity representation makes the entity more distinctive.

        \textbf{Comparison with the SOTA Generative Method. }
        To investigate the comprehensiveness of the advantages of our model, we compare our model with the SOTA generative  model, GEMEL \cite{GEMEL} on accuracy. Smaller values of the quantity of candidate entity $\lambda$ make the task of selecting link entity from $\lambda$ candidates easier.
        Following the configuration of GEMEL, we set $\lambda$ to 16 and the performance is presented in Table \ref{tab:16}. As we can see, our model can still show superiority.
        This is due to our method leveraging cross-modal enhancer and alignment to capture dense interaction over multimodal information while GEMEL mainly relies on the ability of LLM. In fact, the LLM of GEMEL cannot truly learn visual information. Instead, it only depends on CLIP to get visual prefix, inevitably resulting in noise and mismatch between text and image. 


        \begin{table}[tb]
            \centering
            \begin{tabular}{lcc}
            \toprule
                  & WikiMEL & Wikidiverse \\
            \midrule
            GEMEL & 75.5  & 82.7  \\
            DWE(ours)   & 81.0  & 87.1 \\
            \bottomrule
            \end{tabular}%
          \caption{Top-1 accuracy for the model from 16 candidates.}
          \label{tab:16}%
        \end{table}%

\input{Tables/case_study}
    \vspace{-0.3cm}
    \subsection{Ablation Study and Discussion}
        
        \textbf{Attribute enhancement}
        The visual attributes extracted from the image depict the entity's outline and some prominent features. The extracted visual attributes from the image include facial features and adjective-noun pair (ANPs), which play different roles in the model. 
        (1) Facial features: When the entity in the image is a human subject, the extracted facial features serve as a supplement and correction to the feature extractor, such as in cases where the feature extractor fails to recognize a \textit{white male} in the image. As shown in Table \ref{tab:ablation}, by removing the facial features (method 0$\rightarrow$1), the model's performance decreased by 1.6\%, demonstrating the effectiveness of facial features. 
        (2) ANPs: When the image depicts non-human entities like places, ANPs are coarse-grained concepts that describe scenes, such as ``nice clouds," and ``white dress." Due to the presence of noise in ANPs, we use a gated fusion to integrate them. By removing (method 1$\rightarrow$2) attribute-enhanced mention $m_s$ (ANPs), the accuracy decreased by 1.1\%.

        \textbf{Refined visual feature}
        The noise in the image brings challenge to the multimodal model. In this perspective of view, DWE takes visual information refinement to suppress the noise in the image. Table \ref{tab:visual} shows the differences between using refined visual feature $v$ and using primitive visual feature that are extracted from the whole image. It could be seen that the refinement brings 2.1\% improvement over Top-1 accuracy.
        Besides, Tables \ref{tab:visual} that the effectiveness of the refined visual feature varies on datasets. The 8.8\% improvement on WikiMEL(64.0\%$\rightarrow$72.8\%) is more significant than 5.8\% on Richpedia (66.7\%$\rightarrow$72.5\%). The reason is that the image in WikiMEL tends to be noisy with multiple objects, while the object number in Richpedia is less, suggesting that the mechanism is more applicable for noisy datasets (such as WikiMEL). Apart from this, Table \ref{tab:ablation} shows our method without visual feature $v$: the 3.8\% drop(method 3$\rightarrow$4) of Top-1 accuracy shows the importance of vision. 

        \textbf{Crossmodal alignment}
        The 2.5\% decrease of improvement (method2$\rightarrow$3) shows the crossmodal alignment helps to reduce the degree of difference between two features (e.g. partial facial feature $f$ and partial visual feature $d$, text feature $t$ and refined visual feature $v$).

    \subsection{Case Study}
        We further analyze the effectiveness of our model DWE by cases in Table \ref{tab:casestudy}, including correct and wrong predictions. 

        We first analyze the \textbf{correct} predictions. The correctness of case 1 and 4 shows the case with multiple visual objects on different datasets. As for the correct case 3, the mention is \textit{Miller} with two old men in image. The result shows the effectiveness of the attributes. The model could mine the relationship between \textit{two old men}, and entity representation of \textit{Sienna Miller }, while the others are female.
        On the other hand, \textbf{wrong} predictions can still show the effectiveness of our model. In case 2, the visual and textual information includes many objects such as \textit{Khan} and \textit{Khanna}. It is difficult to distinguish these two mentions by the same multimodal information. However, it could be seen that the similarity of \textit{Rajesh Khanna} and \textit{Shah Rukh Khan} is much larger than the rest of candidate entity, which still shows learning ability.
        
        The above cases illustrate the ability of DWE to mine the semantic relationship between textual information (e.g., text and mention) and entity representation. Besides, the supplementary visual feature and attribute extracted from image are leveraged for further disambiguation.
        In summary, the proposed mechanism helps DWE to understand multimodal information and learn joint feature for entity linking.
        
\section{Conclusion}

    In this paper, we explore Multimodal Entity Linking in a neural matching formulation and design a dual-way enhanced framework: query is enhanced by refined multimodal information and the semantics of entity representation is enriched by Wikipedia.
    DWE utilizes a pretrained visual encoder and vision tools to acquire the visual representation and visual attributes, which are subsequently refined into learnable visual characteristics. It regards learnable visual characteristics and textual content as refined information, each contributing to the enhancement of the mention via three enhanced units.
    During the process, we apply the crossmodal alignment to bridge the semantic gap among visual and textual features to gain the multi-view alignment representation.
    Extensive experiments show that DWE outperforms the state-of-the-art methods on three public datasets. Furthermore, we introduce an enhanced entity representation, i.e., textual Wikipedia description. The enhanced entity representation is more representative and no longer restricts the learning of multimodal information.


\clearpage
\section{Acknowledgements}
This work was supported by the National Natural Science Foundation of China (No.72188101, No.62302144).
This work was supported by the Hunan Provincial Natural Science Foundation Project (No.2022JJ30668, 2022JJ30046); and the science and technology innovation program of Hunan province under grant No. 2021GK2001.

\bibliography{AAAI}


\end{document}

%% file: Tables/main.tex
\begin{table*}[htbp]
  \centering
  \footnotesize
    \centering
    \begin{tabular}{l|llll|llll|llll}
    \toprule
    \multicolumn{1}{c|}{\multirow{2}[2]{*}{Models}} & \multicolumn{4}{c|}{\textbf{Richpedia}} & \multicolumn{4}{c|}{\textbf{WikiMEL}} & \multicolumn{4}{c}{\textbf{Wikidiverse}} \\
          & Top-1 & Top-5 & Top-10 & Top-20 & Top-1 & Top-5 & Top-10 & Top-20 & Top-1 & Top-5 & Top-10 & Top-20 \\
    \midrule
    BLINK & 30.8  & 38.8  & 44.5  & 53.6  & 30.8  & 44.6  & 56.7  & 66.4  & -     & 71.2  & -     & - \\
    DZMNED & 29.5  & 41.6  & 45.8  & 55.2  & 30.9  & 50.7  & 56.9  & 65.1  & -     & 39.1  & -     & - \\
    JMEL  & 29.6  & 42.3  & 46.6  & 54.1  & 31.3  & 49.4  & 57.9  & 64.8  & 21.9  & 54.5  & 69.9  & 76.3  \\
    BERT  & 31.6  & 42.0  & 47.6  & 57.3  & 31.7  & 48.8  & 57.8  & 70.3  & 22.2  & 53.8  & 69.8  & 82.8  \\
    ARNN  & 31.2  & 39.3  & 45.9  & 54.5  & 32.0  & 45.8  & 56.6  & 65.0  & 22.4  & 50.5  & 68.4  & 76.6  \\
    MEL-HI & 34.9  & 43.1  & 50.6  & 58.4  & 38.7  & 55.1  & 65.2  & 75.7  & 27.1  & 60.7  & 78.7  & 89.2  \\
    HieCoAtt & 37.2  & 46.8  & 54.2  & 62.4  & 40.5  & 57.6  & 69.6  & 78.6  & 28.4  & 63.5  & 84.0  & 92.6  \\
    GHMFC & 38.7  & 50.9  & 58.5  & 66.7  & 43.6  & 64.0  & 74.4  & 85.8  & -     & -     & -     & - \\
    MMEL  & -     & -     & -     & -     & 71.5  & 91.7  & 96.3  & 98.0  & -     & -     & -     & - \\
    CLIP  & 60.4  & 96.1  & 98.3  & 99.2  & 36.1  & 81.3  & 92.8  & 98.3  & 42.4  & 80.5  & 91.7  & 96.6  \\
    CLIP-text$^\dagger$ & 59.2  & 91.1  & 94.6  & 97.4  & 50.2  & 87.4  & 93.5  & 97.2  & 35.8  & 79.6  & 89.1  & 95.3  \\
    CLIP$^\dagger$ & 63.9  & 96.2  & 98.3  & 99.2  & 58.0  & 95.9  & 98.3  & 99.5  & 43.1  & 80.2  & 90.7  & 96.2  \\
    \midrule\midrule
    DWE & 67.6  & 97.1  & 98.6  & 99.5  & 44.7  & 65.9  & 80.8  & 93.2  & 47.5  & 81.3  & 92.0  & 96.9  \\
    STDEV &  {$\pm$0.74 } & {$\pm$0.11 } & {$\pm$0.07 } & {$\pm$0.05 } & {$\pm$0.08 } & {$\pm$0.15 } & {$\pm$0.18 } & {$\pm$0.20 } & {$\pm$0.11 } & {$\pm$0.15 } & {$\pm$0.29 } & {$\pm$0.23 } \\
    \midrule\midrule
    DWE$^\dagger$ & \textbf{72.5 } & \textbf{97.3 } & \textbf{98.8 } & \textbf{99.6 } & \textbf{72.8 } & \textbf{97.5 } & \textbf{98.9 } & \textbf{99.7 } & \textbf{51.2 } & \textbf{91.0 } & \textbf{96.3 } & \textbf{98.9 } \\
    STDEV & {$\pm$0.28 } & {0.06 } & {$\pm$0.05 } & {$\pm$0.07 } & {$\pm$0.63 } & {$\pm$0.06 } & {$\pm$0.05 } & {$\pm$0.05 } & {$\pm$1.04 } & {$\pm$0.23 } & {$\pm$0.11 } & {$\pm$0.07 } \\
    \bottomrule
    \end{tabular}%
    
    \caption{MEL results of the compared models in terms of Top-1, 5, 10, 20 accuracy (\%) from $\lambda$ candidate entites, where $\lambda=100$.
    For the model, the results with $\dagger$ are using enhanced entity representations (textual Wikipedia demonstration), whereas the others represent the results using previous entity representation.
    We not only report the best performance but also the standard deviation (STDEV) over 10 runs with random initialization. The best results are highlighted in bold.
    }
    \label{tab:experiment}
\end{table*}%

%% file: Tables/ablation.tex

\begin{table}[tb]
  \centering
    \begin{tabular}{p{0.1cm}p{0.1cm}p{0.1cm}p{0.1cm}p{0.1cm}p{0.1cm}C{0.8cm}C{0.8cm}C{0.9cm}C{0.9cm}}
    \toprule
      &   $m_t$     & $m_v$     & $\mathcal{L}_c $  & $m_s$  &  $f$  & Top-1 & Top-5 & Top-10 & Top-20 \\
    \midrule
    \textbf{0} &  \ding{51}     & \ding{51}     & \ding{51}     & \ding{51}     & \ding{51}     & 72.5  & 97.3  & 98.8  & 99.6 \\
    1     &  \ding{51}     & \ding{51}     & \ding{51}     & \ding{51}     & \ding{55}     & 70.9  & 97.5  & 98.8  & 99.5  \\
    2     &  \ding{51}     & \ding{51}     & \ding{51}     & \ding{55}     & \ding{55}     & 69.8  & 97.3  & 98.4  & 99.4  \\
    3     &  \ding{51}     & \ding{51}     & \ding{55}     & \ding{55}     & \ding{55}     & 67.3  & 93.9  & 98.4  & 99.4  \\
    4     &  \ding{51}     & \ding{55}     & \ding{55}     & \ding{55}     & \ding{55}     & 63.5  & 92.7  & 96.0  & 98.3  \\
    \bottomrule
    \end{tabular}%
  \caption{Ablation study on Richpedia ($m_t$: textual enhanced feature; $m_v$: visual enhanced feature; $\mathcal{L}_c $: crossmodal alignment; $m_s$: attribute-enhanced mention; $f$: facial characteristics). Method 0 represents our model.}
  \label{tab:ablation}%
\end{table}%

\begin{table}[tb]
  \centering
    \begin{tabular}{lcccc}
    \toprule
    Features & Top-1 & Top-5 & Top-10 & Top-20 \\
    \midrule
    primitive on rich & 66.7  & 97.2  & 98.8  & 99.6  \\
    refined on rich & 72.5  & 97.3  & 98.8  & 99.6 \\
    \midrule\midrule
    primitive on wiki & 64.0  & 96.2  & 98.4  & 99.5  \\
    refined on wiki & 72.8  & 97.5  & 98.9  & 99.7 \\
    \bottomrule
    \end{tabular}%
  \caption{Ablation study on using refined visual feature $v$ or primitive visual feature which is extracted directly from image over public datasets: Richpedia (rich), WikiMEL (wiki)}
  \label{tab:visual}%
\end{table}%

%% file: Tables/case_study.tex
\begin{table*}[pbt]
    \renewcommand\arraystretch{0.83}
    \small
    \begin{tabular}{p{0.8cm}p{3.75cm}p{3.75cm}p{3.75cm}p{3.75cm}}
        \toprule
        Case    & \makebox[2.5cm][c]{1}    & \makebox[2.5cm][c]{2} & \makebox[2.5cm][c]{3} & \makebox[2.5cm][c]{4} \\
        \midrule
        Image
                & \makebox[3cm][c]{\begin{minipage}[c]{0.24\columnwidth}\centering{\includegraphics[width=\textwidth]{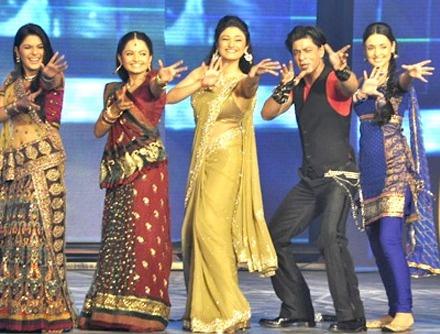}}\end{minipage} }
                & \makebox[3cm][c]{\begin{minipage}[c]{0.24\columnwidth}\centering{\includegraphics[width=\textwidth]{utils/cs1_1008.jpg}}\end{minipage} }
                & \makebox[3cm][c]{\begin{minipage}[c]{0.29\columnwidth}\centering{\includegraphics[width=\textwidth]{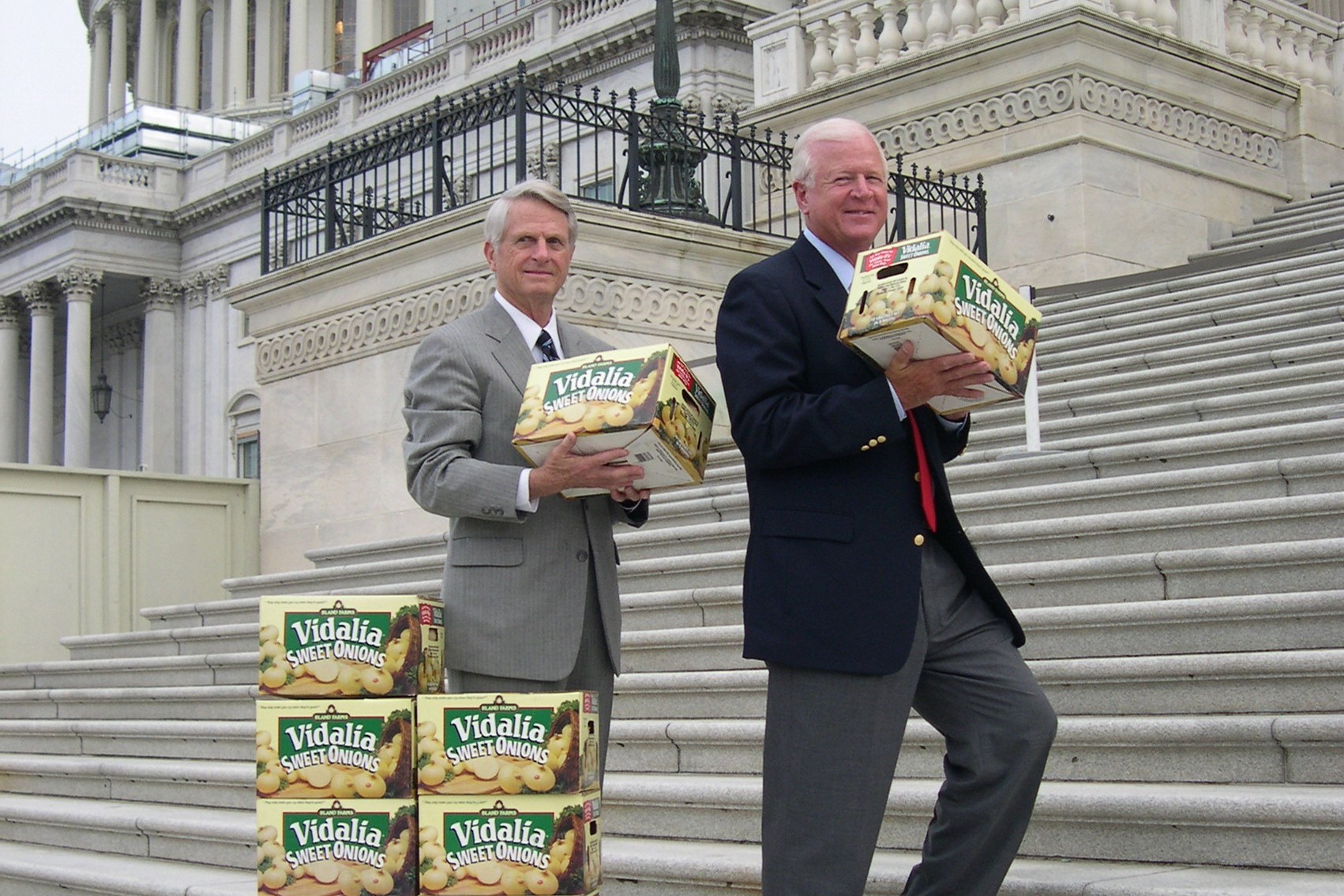}}\end{minipage} }
                & \makebox[3cm][c]{\begin{minipage}[c]{0.22\columnwidth}\centering{\includegraphics[height=\textwidth]{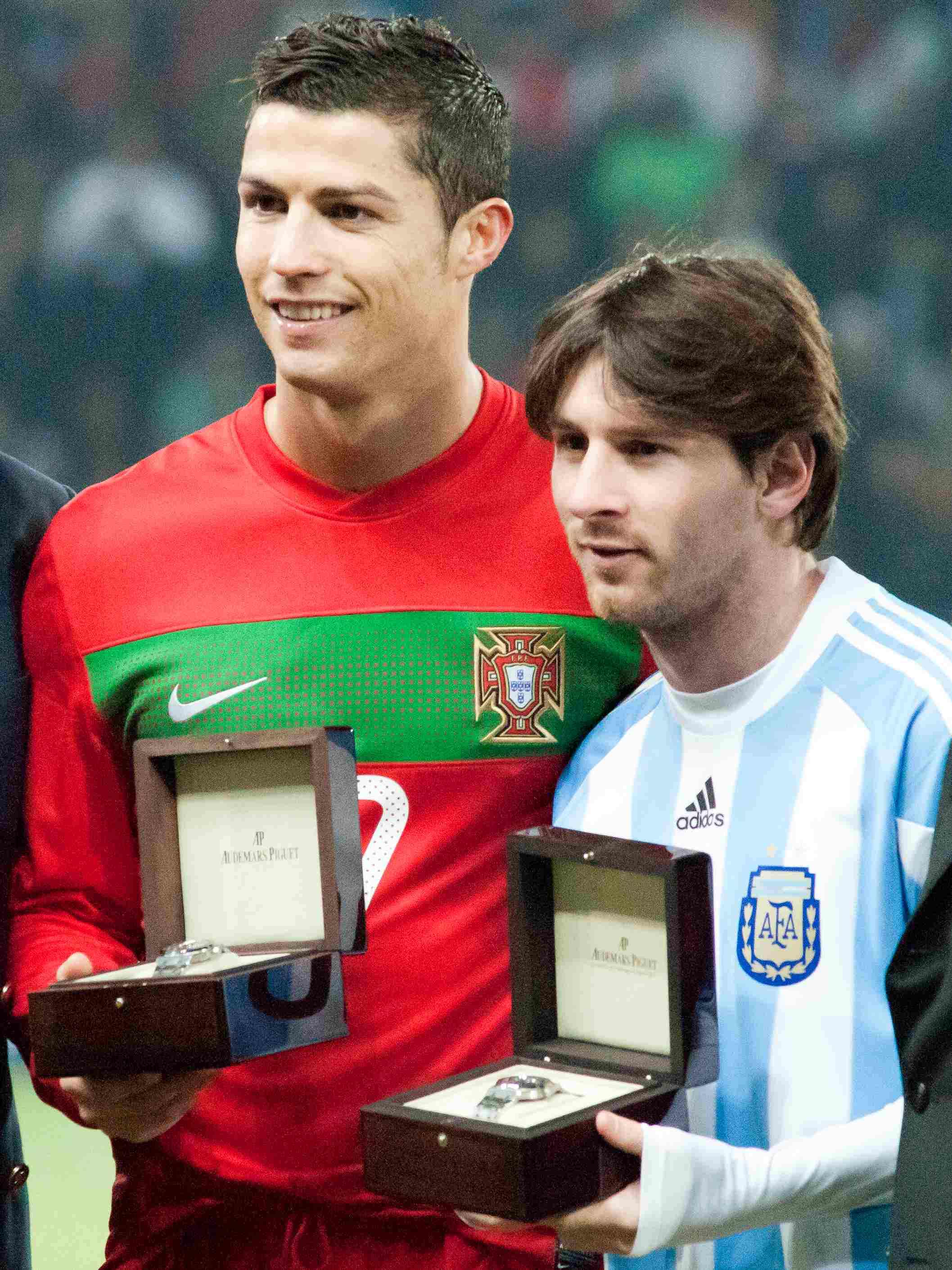}}\end{minipage} }                                                                                                                                                           \\
        Text
                & Khanna and \textbf{Manek} at the music launch of Ra.One 
                & \textbf{Khanna} and Manek at the music launch of Ra.One
                & \textbf{Miller} and Saxby Chambliss in 2004
                & \textbf{Messi} has scored over 700 senior career goals                                                                                                                                                                                                   \\
        \hline
        Mention & \makebox[3cm][c]{Manek}    & \makebox[3cm][c]{Khan}                & \makebox[3cm][c]{Miller}                & \makebox[3cm][c]{Messi}               \\
        Entity
                & \textit{Giaa Manek} is an Indian television actress, who played a role in film Na Ghar Ke..
                & \textit{Shah Rukh Khan} also known by the initialism SRK, is an Indian actor, film producer..
                & \textit{Zell Bryan Miller} (1932-2013) was an American author and politician..
                & \textit{ Lionel Messi}, also known as Leo Messi, is an Argentine professional footballer..                                                                                         \\
        \tabincell{l}{Top-5 \\ candidate}
                & \tabincell{l}{{\textbf{Giaa Manek 0.91}} \\ Gia Kancheli 0.34				\\ Mihir Bipin Manek 0.29	\\ Brady Manek 0.28					\\ Rama 0.04			}
                & \tabincell{l}{Rajesh Khanna 0.76 \\ {\textbf{Shah Rukh Khan 0.72}}		\\ Zareen Khan 0.43			\\ Gauhar Khan 0.33					\\ Aamir Khan 0.19		}
                & \tabincell{l}{{\textbf{Sienna Miller 0.79}} \\ Marisa Miller 0.61				\\ Kristine Miller 0.56		\\ Ezra Miller 0.31					\\ Dean Miller 0.30		}
                & \tabincell{l}{{\textbf{Lionel Messi 0.85}} \\ Antonello da Messina 0.76		\\ Charles Messier 0.56		\\ Massiv	0.29 					\\ Ramesses II 0.06		}      \\
        Predict  & \makebox[3cm][c]{{\ding{51}}}  & \makebox[3cm][c]{{\ding{55}}} & \makebox[3cm][c]{{\ding{51}}} & \makebox[3cm][c]{{\ding{51}}} \\
        \bottomrule
    \end{tabular}
    \caption{Multimodal entity linking cases. Entity is depicted by textual Wikipedia description. Top-5 entities are selected from $\lambda$ candidate entities. We take the bold text to show the correct entity in Top-5 candidate. Case 1 and 2 are multiple object samples for different mentions with the same image and text.}
    \label{tab:casestudy}
\end{table*}